# Filtering for More Accurate Dense Tissue Segmentation in Digitized Mammograms

Mario Mustra, Mislav Grgic

University of Zagreb, Faculty of Electrical Engineering and Computing, Unska 3, 10000 Zagreb, Croatia
*mario.mustra@fer.hr*

*Abstract* - **Breast tissue segmentation into dense and fat tissue is important for determining the breast density in mammograms. Knowing the breast density is important both in diagnostic and computer-aided detection applications. There are many different ways to express the density of a breast and good quality segmentation should provide the possibility to perform accurate classification no matter which classification rule is being used. Knowing the right breast density and having the knowledge of changes in the breast density could give a hint of a process which started to happen within a patient. Mammograms generally suffer from a problem of different tissue overlapping which results in the possibility of inaccurate detection of tissue types. Fibroglandular tissue presents rather high attenuation of X-rays and is visible as brighter in the resulting image but overlapping fibrous tissue and blood vessels could easily be replaced with fibroglandular tissue in automatic segmentation algorithms. Small blood vessels and microcalcifications are also shown as bright objects with similar intensities as dense tissue but do have some properties which makes possible to suppress them from the final results. In this paper we try to divide dense and fat tissue by suppressing the scattered structures which do not represent glandular or dense tissue in order to divide mammograms more accurately in the two major tissue types. For suppressing blood vessels and microcalcifications we have used Gabor filters of different size and orientation and a combination of morphological operations on filtered image with enhanced contrast.**

## I. Introduction

Computer-aided diagnosis (CAD) systems have become an integral part of modern medical systems and their development should provide a more accurate diagnosis in shorter time. Their aim is to help radiologists, especially in screening examinations, when a large number of patients are examined and radiologist often spend very short time for readings of non-critical mammograms. Mammograms are X-ray breast images of usually high resolution with moderate to high bit-depth which makes them suitable for capturing fine details. Because of the large number of captured details, computer-aided detection (CADe) systems have difficulties in detection of desired microcalcifications and lesions in the image. Since mammograms are projection images in grayscale, it is difficult to automatically differentiate types of breast tissue because different tissue types can have the same or very similar intensity. The problem which occurs in different mammograms is that the same tissue type is shown with a different intensity and therefore it is almost impossible to set an accurate threshold based solely on the histogram of the image. To overcome that problem different authors have came up with different solutions. Some authors used statistical feature extraction and classification of mammograms into different categories according to their density. Other approaches used filtering of images and then extracting features from filter response images to try getting a good texture descriptor which can be classified easier and provide a good class separability. Among methods which use statistical feature extraction, Oliver et al. [1] obtained very good results using combination of statistical features extracted not directly from the image, but from the gray level co-occurrence matrices. We can say that they have used $2^{nd}$ order statistics because they wanted to describe how much adjacent dense tissue exists in each mammogram. Images were later classified into four different density categories, according to BI-RADS [2]. Authors who used image filtering techniques tried to divide, as precisely as possible, breast tissue into dense and fat tissue. With the accurate division of dense and fat tissue in breasts it would be possible to quantify the results of breast density classification and classification itself would become trivial because we would have numerical result in number of pixels belonging to the each of two groups. However, the task of defining the appropriate threshold for dividing breast tissue into two categories is far from simple. Each different mammogram captured using the same mammography device is being captured with slightly different parameters which will affect the final intensity of the corresponding tissues in the different image. These parameters are also influenced by the physical property of each different breast, for example its size and amount of dense and fat tissue. In image acquisition process the main objective is to produce an image with very good contrast and no clipping in both low and high intensity regions. Reasons for different intensities for the corresponding tissue, if we neglect usage of different imaging equipment, are difference in the actual breast size, difference in the compressing force applied to the breast during capturing process, different exposure time and different anode current. Having this in mind authors tried to overcome that problem by applying different techniques which should minimize influence of capturing inconsistencies. Muhimmah and Zwiggelaar [3] presented an approach of multiscale histogram analysis having in mind that image resizing will affect the histogram shape because of detail removal when image is being downsized. In this way they were able to remove small bright objects from images and tried to get satisfactory results by determining which objects correspond to large tissue areas. Petroudi et al. [4] used Maximum Response 8 filters [5] to obtain a texton dictionary which was used in conjunction with the support-vector machine classifier to classify the breast into four density categories. Different equipment for capturing mammograms produces resulting images which have very different properties. The most common division is in two main categories: SFM (Screen Film Mammography) and FFDM (Full-Field Digital Mammography). Tortajada et al. [6] have presented a work in which they try to compare accuracy of the same classification method on the SFM and the FFDM images. Results which they have obtained show that there is a high correlation of automatic classification and expert readings and overall results are slightly better for FFDM images. Even inside the same category, such is FFDM, captured images can be very different. DICOM standard [7] recommends bit-depth of 12 bits for mammography modality and images are stored according to the recommendation. However if we later observe the histogram of those images, it is clear that the actual bit-depth is often much lower and is usually below 10 bits.

In this paper we present a method which should provide a possibility for division of the breast tissue between parenchymal tissue and fatty tissue without influence of fibrous tissue, blood vessels and fine textural objects which surround the fibroglandular disc. Segmentation of dense or glandular tissue from the entire tissue will be made by setting different thresholds. Our goal is to remove tissue which interferes with dense tissue and makes the division less accurate because non-dense tissue is being treated as dense due to its high intensity when compared to the rest of the tissue. Gabor filters generally proved to be efficient in extracting features for the breast cancer detection from mammograms because of their sensitivity to edges in different orientations [8]. Therefore, for the removal of blood vessels, we have used Gabor filter bank which is sensitive to brighter objects which are rather narrow or have high spatial frequency. Output of the entire filter bank is an image which is created from superimposed filter responses of different orientations. Subtraction of the image which represents vessels and dense tissue boundaries from the original image produces a much cleaner image which can later be enhanced in order to equalize intensity levels of the corresponding tissue types among different images. In that way we will be able to distinct dense tissue from fat more accurately. The proposed method has been tested on mammograms from the mini-MIAS database [9] which are all digitized SFMs.

This paper is organized as follows. In Section II we present the idea behind image filtering using Gabor filter bank and explain which setup we will use for filtering out blood vessels and smaller objects. In Section III we present results of filtering with the appropriate filter and discuss results of region growing after contrast enhancement and application of morphological operations. Section IV draws the conclusions.

## II. GABOR FILTERS

Gabor filters are linear filters which are most commonly used for the edge detection purposes as well as the textural feature extraction. Each filter can be differently constructed and it can vary in frequency, orientation and scale. Because of that Gabor filters provide a good flexibility and orientation invariantism. Gabor filter in a complex notation can be expressed as:

$$G = \exp\left(-\frac{(x\cos\theta + y\sin\theta)^2 + \gamma^2(y\cos\theta - x\sin\theta)^2}{2\sigma^2}\right) \cdot \exp\left(i\left(\frac{2\pi(x\cos\theta + y\sin\theta)}{\lambda} + \psi\right)\right), \quad (1)$$

where $\theta$ is the orientation of the filter, $\gamma$ is the spatial aspect ratio, $\lambda$ is the wavelength of the sinusoidal factor, $\sigma$ is the sigma or width of the Gaussian envelope and $\psi$ is the phase offset. This gives a good possibility to create different filters which are sensitive to different objects in images. To be able to cover all possible blood vessels and small linearly shaped objects it is necessary to use more than one orientation. In our experiment we have used 8 different orientations and therefore obtained the angle resolution of 22.5°. Figure 1 (a)-(h) shows 8 different filter orientations created using (1) with the angle resolution of 22.5° between each filter, from 0 to 157.5° respectively.

Besides the orientation angle, one of the most commonly changed variables in (1) is the sinusoidal frequency. Different sinusoidal frequencies will provide different sensitivity of the used filter for different spatial frequencies of objects in images. If the chosen filter contains more wavelengths, filtered image will correspond more to the original image because filters will be sensitive to objects of a high spatial frequency, e.g. details. In the case of smaller number of wavelengths, filtered image will contain highly visible edges. Figure 2 shows different wavelengths of Gabor filter with the same orientation.

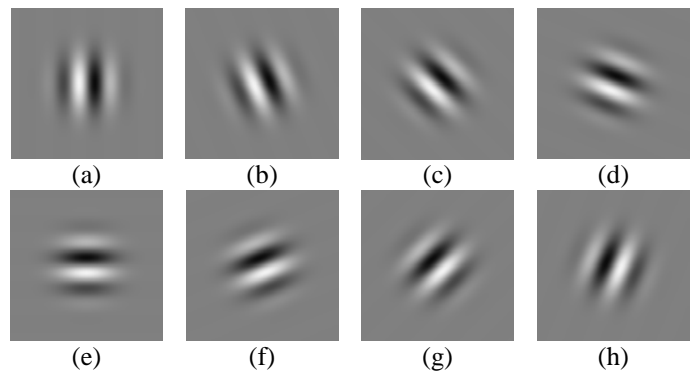

Figure 1. Gabor filters of the same scale and wavelength with different orientations: (a) 0°; (b) 22.5°; (c) 45°; (d) 67.5°; (e) 90°; (f) 112.5°; (g) 135°; (h) 157.5°.

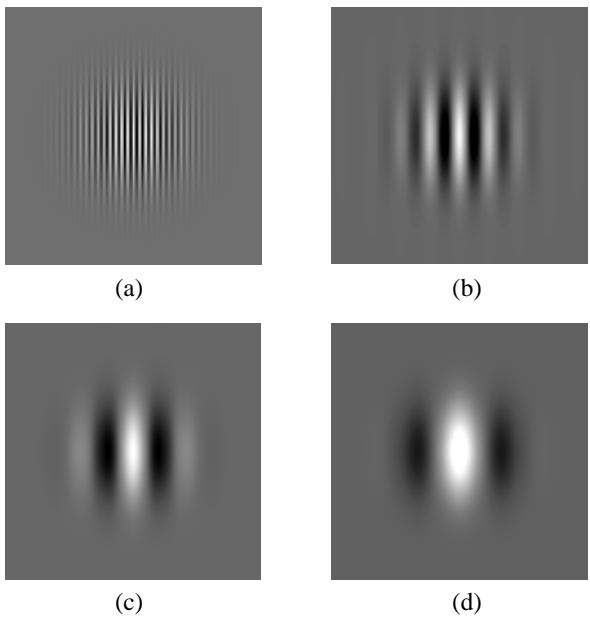

Figure 2. (a)-(d) Gabor filters which contains larger to smaller number of sinusoidal wavelengths respectively.

There is of course another aspect of the filter which needs to be observed and that is the actual dimension of the filter. Dimension of the filter should be chosen carefully according to image size and the size of object which we want to filter out and should be less than 1/10 of the image size.

### III. IMAGE FILTERING

Preprocessing of images is the first step which needs to be performed before filtering. Preprocessing steps include image registration, background suppression with the removal of artifacts and the pectoral muscle removal in the case of Medio-Lateral Oblique mammograms. For this step we have used manually drawn segmentation masks for all images in the mini-MIAS database. These masks were hand-drawn by an experienced radiologist and, because of their accuracy, can be treated as ground truth since there is no segmentation error which can occur as the output of automatic segmentation algorithms. The entire automatic mask extraction process has been described in [10] and steps for the image "mdb002" from the mini-MIAS database are shown in Figure 3.

After the preprocessing we have proceeded with locating of the fibroglandular disc position in the each breast image. Fibroglandular disc is a region containing mainly two tissue types, dense or glandular and fat and according to their distribution it is possible to determine in which category according to the density some breast belong.

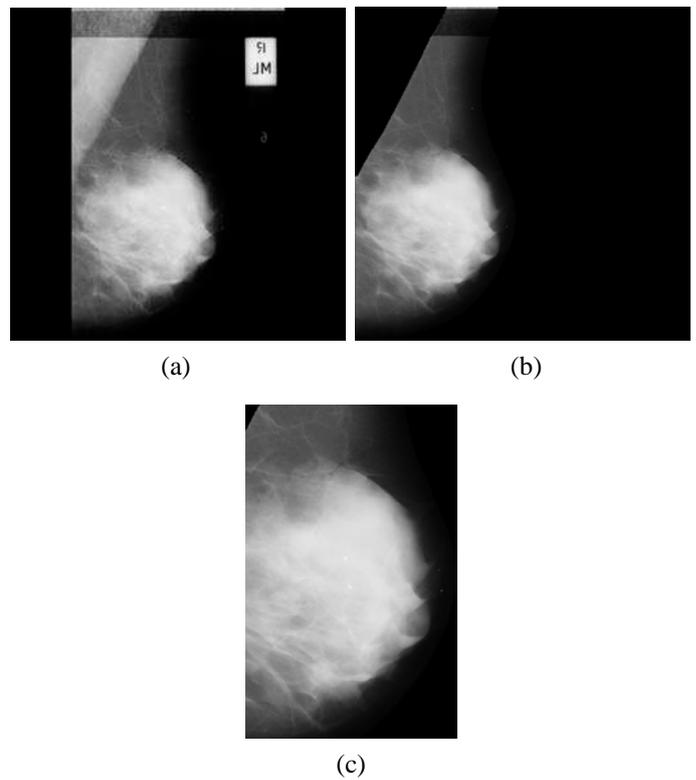

Figure 3. (a) Original mini-MIAS image "mdb002"; (b) Registered image with removed background and pectoral muscle; (c) Extracted ROI from the same image which will be filtered using Gabor filter.

Dense tissue mainly has higher intensity in mammograms because it presents higher attenuation for X-rays than fat tissue. Intensity also changes with the relative position towards edge of the breast because of the change in thickness. Since the fibroglandular disc is our region of interest, we have extracted only that part of the image. Entire preprocessing step done for all images in the mini-MIAS database is described in [11]. To extract ROI we have cropped part of the image according to the maximum breast tissue dimensions as shown in Figure 4. Actual ROI boundaries are chosen to be V and H for vertical and horizontal coordinates respectively:

$$V = \left[ \frac{\max(horizontal)}{2} : \frac{\max(horizontal)}{2} + \max(horizontal) \right]$$
$$H = \left[ \frac{\max(vertical)}{3} : \max(vertical) \right] \quad (2)$$

where max(*horizontal*) is the vertical coordinate of the maximal horizontal dimension, and max(*vertical*) is the horizontal coordinate of the maximal vertical dimension.

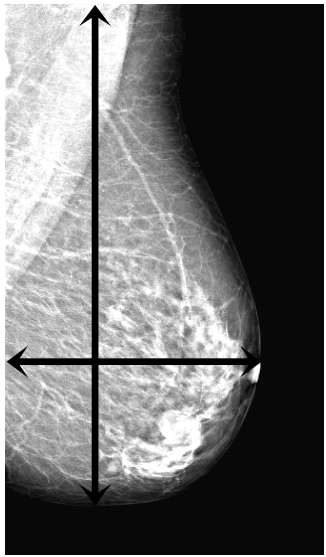

Figure 4. Defining maximal breast tissue dimension for ROI extraction.

With this approach we get a large part or entire fibroglandular disc isolated and there is no need for the exact segmentation of it. It would be good if we could eliminate fibrous tissue and blood vessels and treat our ROI as it is completely uniform in the case of low density breasts. To be able to perform that task we can choose an appropriate Gabor filter sensitive to objects that we want to remove. A good Gabor filter for detection of objects with high spatial frequency contains less sinusoidal wavelengths, like the ones showed in Figure 2 (c) and (d).

Contrast Limited Adaptive Histogram Equalization (CLAHE) [12] is a method for local contrast enhancement which is suitable for equalization of intensities in each ROI that we observe. Contrast enhancement obtained using CLAHE method will provide better intensity difference between dense and fat tissue. CLAHE method is based on a division of the image into smaller blocks to allow better contrast enhancement and at the same time uses clipping of maximal histogram components to achieve a better enhancement of mid-gray components. If we observe the same ROI before and after applying CLAHE enhancement it is clear that fat tissue can be filtered out easier after the contrast enhancement. Figure 5 shows application of the contrast enhancement using CLAHE on "mdb001".

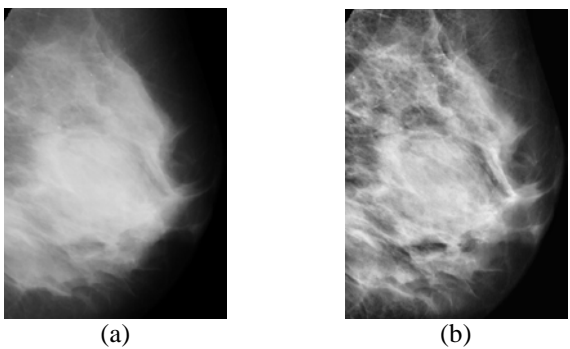

Figure 5. (a) Original ROI from "mdb001"; (b) Same ROI after the contrast enhancement using CLAHE.

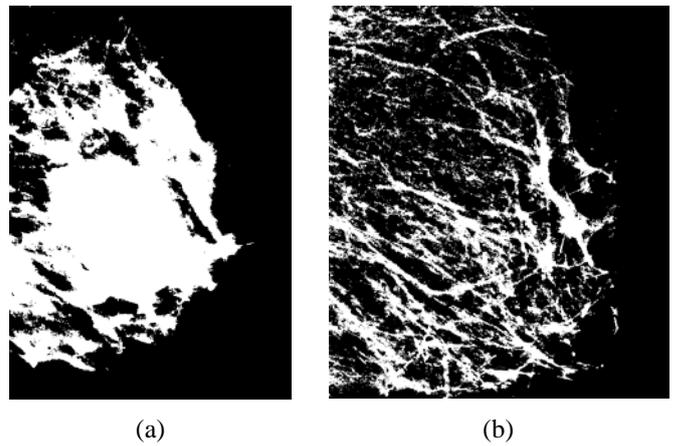

Figure 6. (a) Enhanced ROI from "mdb001" with the applied threshold; (b) Enhanced ROI from "mdb006" with the applied threshold.

If we apply the threshold on the enhanced ROI we will get the result for "mdb001" and "mdb006" as shown in Figure 6 (a) and (b). These two images belong to the opposite categories according to the amount of dense tissue. The applied threshold is set to 60% of the mean image intensity.

After applying threshold, images have visibly different properties according to the tissue type. It is not possible to apply the same threshold because different tissue type has different intensities. Contrast enhancement makes the detection of fibrous tissue and vessels easier especially after Gabor filtering, Figure 7 (a) and (b).

After contrast enhancement and filtering images using Gabor filter to remove fibrous tissue we need to make a decision in which category according to density each breast belongs. For that we will use binary logic with different threshold applied to images. We will apply two thresholds, at 60% and 80% of the maximal intensity and calculate the area contained in both situations. For that we will use logical AND operator, Figure 8. From Figure 8 (e) and (f) we can see that combination of threshold images using the logical AND for low and high density give the correct solution. Figures 8 (e) and (f) show the result of (a) AND (c) and (b) AND (d).

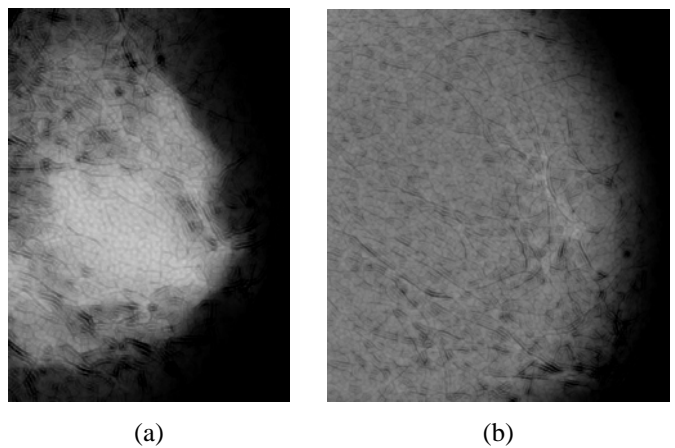

Figure 7. (a) ROI from "mdb001" after applying Gabor filter; (b) ROI from "mdb006" after applying Gabor filter.

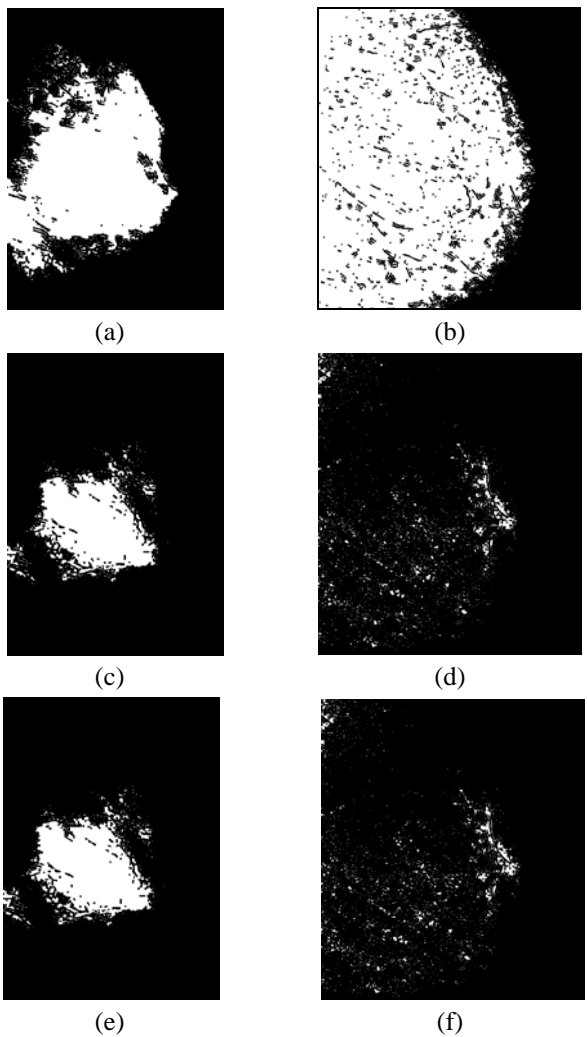

Figure 8. (a) "mdb001" after filtering out blood vessels and threshold at 60% of maximal intensity; (b) "mdb006" after filtering out blood vessels and threshold at 60% of maximal intensity; (c) "mdb001", threshold at 80%; (d) "mdb006", threshold at 80%; (e) "mdb001" threshold at 60% AND threshold at 80%; (f) "mdb006" threshold at 60% AND threshold at 80%.

IV. CONCLUSION

In this paper we have presented a method which combines the usage of local contrast enhancement biased on the mid-gray intensities. The contrast enhancement has been achieved using CLAHE method. Gabor filters have been used for the removal of blood vessels and smaller portions of fibrous tissue which has similar intensity as dense tissue. Combination of different thresholds in conjunction with the logical AND operator provided a good setup for determining whether we have segmented a fat or dense tissue and therefore can give as a rule for segmenting each mammogram with different threshold, no matter what the overall tissue intensity is. The advantage of Gabor filter over classical edge detectors is in easy orientation changing and possibility to cover all possible orientations by superpositioning filter responses. Usage of Gabor filters improves the number of false positive results which come from blood vessels or small fibrous tissue segments and contrast enhancement provides a comparability of the same tissue type in different mammograms. Our future work in this field will be development of automatic segmentation algorithms for dense tissue in order to achieve a quantitative breast density classification by knowing the exact amount of the dense tissue.

ACKNOWLEDGMENT

The work described in this paper was conducted under the research project "Intelligent Image Features Extraction in Knowledge Discovery Systems" (036-0982560-1643), supported by the Ministry of Science, Education and Sports of the Republic of Croatia.